\documentclass{article}

\usepackage{times}
\usepackage{graphicx} 
\usepackage{subfigure}

\usepackage{array}
\usepackage[T1]{fontenc}

\usepackage{natbib}

\usepackage{algorithm}
\usepackage{algorithmic}

\usepackage[english]{babel}
\usepackage{amsmath,amsfonts,amsthm,amssymb}

\usepackage{hyperref}

\usepackage[accepted]{icml2016}

\usepackage{url}

\icmltitlerunning{Learning to Generate with Memory}

\newcommand{\bbtheta}{\boldsymbol{\theta}}

\newcommand{\bbmu}{\boldsymbol{\mu}}

\newcommand{\bbx}{\mathbf{x}}

\newcommand{\bbh}{\mathbf{h}}

\newcommand{\bbz}{\mathbf{z}}

\newcommand{\junx}[1]{{\color{red}{\bf\sf [JZ: #1]}}}

\begin{document}

\twocolumn[
\icmltitle{Learning to Generate with Memory}

\icmlauthor{Chongxuan Li}{licx14@mails.tsinghua.edu.cn}
\icmlauthor{Jun Zhu}{dcszj@mail.tsinghua.edu.cn}
\icmlauthor{Bo Zhang}{dcszb@mail.tsinghua.edu.cn}
\icmladdress{Dept. of Comp. Sci. $\&$ Tech., State Key Lab of Intell. Tech. $\&$ Sys., TNList Lab, \\
Center for Bio-Inspired Computing Research, Tsinghua University, Beijing, 100084, China}

\icmlkeywords{}

\vskip 0.3in
]

\begin{abstract}
Memory units have been widely used to enrich the capabilities of deep networks on capturing long-term dependencies in reasoning and prediction tasks, but little investigation exists on deep generative models (DGMs) which are good at inferring high-level invariant representations from unlabeled data. This paper presents a deep generative model with a possibly large external memory and an attention mechanism to capture the local detail information that is often lost in the bottom-up abstraction process in representation learning. By adopting a smooth attention model, the whole network is trained end-to-end by optimizing a variational bound of data likelihood via auto-encoding variational Bayesian methods, where an asymmetric recognition network is learnt jointly to infer high-level invariant representations. The asymmetric architecture can reduce the competition between bottom-up invariant feature extraction and top-down generation of instance details. Our experiments on several datasets demonstrate that memory can significantly boost the performance of DGMs on various tasks, including density estimation, image generation, and missing value imputation, and DGMs with memory can achieve state-of-the-art quantitative results.
\end{abstract}

\section{Introduction}

Deep learning models are able to extract abstract representations from low-level inputs by adopting a deep architecture with explicitly designed nonlinear transformations~\cite{bengio13}. Among many types of deep models, deep generative models (DGMs) learn abstract representations from unlabeled data and can perform a wide range of tasks, including density estimation, data generation and missing value imputation. Depending on the building blocks, various types of DGMs exist, including undirected models~\cite{Salakhutdinov:09}, directed models~\cite{neal92,hinton:06}, autoregressive models~\cite{Larochelle:11,Gregor:14}, and Markov chain based models~\cite{bengio14}.
Recently, DGMs have attracted much attention on developing efficient and (approximately) accurate learning algorithms, such as stochastic variational methods~\cite{kingma14iclr,danilo14icml,bornschein14,burda15} and Monte Carlo methods~\cite{adams10,gan15,chao15}.

Although current DGMs are able to extract high-level abstract representations, they may not be sufficient in generating high-quality input samples. This is because more abstract representations are generally {\it invariant} or less sensitive to most specific types of local changes of the input. This bottom-up abstraction progress is good for identifying predictive patterns, especially when a discriminative objective is optimized~\cite{Li15}; but it also loses the detail information that is necessary in the top-down generating process.
It remains a challenge for DGMs to generate real data, especially for images that have complex structures. Simply increasing the model size is apparently not wise, as it may lead to serious over-fitting without proper regularization as well as heavy computation burden.
Some recent progress has been made to improve the generation quality.
For example,
DRAW~\cite{gregor15} iteratively constructs complex images over time through a recurrent encoder and decoder together with an attention mechanism and
LAPGAN~\cite{denton15} employs a cascade of generative adversarial networks (GANs)~\cite{goodfellow:14} to generate high quality natural images through a Laplacian pyramid framework~\cite{Burt83}.
However, no efforts have been made on enriching the capabilities of probabilistic DGMs by designing novel building blocks in the generative model. 

In this paper, we address the above challenges by presenting a new architecture for building probabilistic deep generative models with a possibly large external memory and an attention mechanism. Although memory has been explored in various deep models for capturing long-term dependencies in reasoning and prediction tasks (See Section~\ref{sec:related-work} for a review), our work represents a first attempt to leverage external memory to enrich the capabilities of probabilistic DGMs for better density estimation, data generation and missing value imputation. The overall architecture of our model is an interleave between stochastic layers and deterministic layers, where each deterministic layer is associated with an external memory to capture local variant information. An attention mechanism is used to record information in the memory during learning and retrieve information from the memory during data generation.
This attention mechanism can be trained because the invariant information and local variant information are correlated, e.g., both containing implicit label information.
Both the memory and attention mechanisms are parameterized as differentiable components with some smooth nonlinear transformation functions. Such a design allows us to learn the whole network end-to-end by developing a stochastic variational method, which introduces a recognition network without memory to characterize the variational distribution.
Different from~\cite{kingma14iclr,burda15}, our recognition network is asymmetric to the generative network. This asymmetric recognition network is sufficient for extracting invariant representations in bottom-up inference, and is compact in parameterization.
Furthermore, this asymmetry can help reduce the competition between bottom-up invariant feature extraction (using the recognition network) and top-down input generation (using the deep generative model with memory).

We quantitatively and qualitatively evaluate our method on several datasets in various tasks, including density estimation, data generation and missing value imputation. Our results demonstrate that an external memory together with a proper attention mechanism can significantly improve DGMs to obtain state-of-the-art performance.


\section{Related Work}
\label{sec:related-work}

Memory has recently been leveraged in deep models to capture long-term dependencies for various tasks, such as algorithm inference~\cite{graves14},
question answering~\cite{weston15,suknbaatar15} and neural language transduction~\cite{grefenstette15}.
The external memory in these models provides a way to record information stably and interact with the environment,
and hence extends the capability of traditional learning models.
Typically the interaction, e.g., reading from and writing on the memory, is done through an associated attention mechanism and the whole system is trained with supervision.
The attention mechanism can be differentiable and trained in an end-to-end manner~\cite{graves14,suknbaatar15},
or really discrete and trained by a Reinforcement Learning algorithm~\cite{Zaremba15}.

In addition to the memory-based models mentioned above,
attention mechanisms have been used in other deep models for various tasks,
such as image classification~\cite{Larochelle10,Ba15},
object tracking~\cite{Mnih14},
conditional caption generation~\cite{xu15},
machine translation~\cite{Bahdanau15}
and image generation~\cite{graves13,gregor15}.
Recently,
DRAW~\cite{gregor15} introduces a novel 2-D attention mechanism to decide ``where to read and write'' on the image and does well in generating objects with clear track, such as handwritten digits and sequences of real digits.

Compared with previous memory-based networks~\cite{graves14,weston15}, we propose to employ an external hierarchical memory to capture variant information at different abstraction levels trained in an unsupervised manner. Besides, our memory cannot be written directly like~\cite{graves14,weston15}; instead it is updated through optimization.
Compared with previous DGMs with visual attention~\cite{Tang14,gregor15},
we make different assumptions about the data, i.e., the main object (such as faces) has massive local features, which cannot be modeled by a limited number of latent factors. We employ an external memory to capture this and the associated attention mechanism is used to retrieve the memory, not to learn ``what-where'' combination on the images.
Besides, the external memory used in our model and the memory units of LSTMs used in DRAW~\cite{gregor15} can complement each other~\cite{graves14}. Further investigation on DRAW with external memory is our future work.

Considering the bottom-up inference procedure and top-down generation procedure together, additional memory mechanisms can help to reduce the competition between invariant feature extraction and local variant reconstruction, especially when label information is provided (e.g., in supervised or semi-supervised setting). Similar idea is highlighted in the Ladder Network~\cite{Valpola14,rasmus15},
which reconstructs the input hierarchically using an extension of denoising autoencoders (dAEs)~\cite{Vincent10} with the help of lateral connections and achieves excellent performance on semi-supervised learning~\cite{rasmus15}.
Though it is possible to interpret the Ladder Network probabilistically as in~\cite{bengio13nips}, we model the data likelihood directly with the help of external memory instead of explicit lateral edges.
Our method can also be extended to do supervised or semi-supervised learning as in~\cite{kingma14nips}, which is our future work. 

\section{Probabilistic DGMs with Memory}

We present a probabilistic deep generative model (DGM) with a possibly large external memory as well as a soft attention mechanism.

\subsection{Overall Architecture}
\label{sec:overall-struct}

Formally, given a set of training data $\mathcal{D}$, we assume each $\bbx \in \mathcal{D}$ is independently generated with a set of hierarchically organized latent factors $\bbz_L , \dots, \bbz_1$ as follows:
\begin{itemize}\vspace{-.3cm}
\item Draw the top-layer factors $\bbz_L \sim  \mathcal{N}(0, I)$.\vspace{-.3cm}
\item For $l = L-1, \dots , 0$, calculate the mean parameters $\bbmu_l = g_{l}(\bbz_{l+1}; M_{l})$ and draw the factors $\bbz_{l} \sim  P_{l}(\bbmu_l)$,\vspace{-.3cm}
\end{itemize}
where each $g_{l}$ is a nonlinear function, often assumed to be smooth for the ease of learning. To connect with observations, the bottom layer is clamped at $\bbz_0 = \bbx$.
Each $\bbz_l$ is randomly sampled from a Gaussian distribution except $\bbz_0$ whose distribution depends on the properties of the data (e.g., Gaussian for continuous data or Multinomial for discrete case). All the distributions $P_l$ are assumed to be of an exponential family form, with mean parameters $\bbmu_l$. 

Here, we define $g_l$ as a feed-forward deep neural network with $I_l$ deterministic layers and a set of associated memories $\{ M_{l}^{(i)} \}_{i=0}^{I_l - 1}$, one per layer. We parameterize each memory as a trainable matrix with dimension $d_s \times n_s$, where $d_s$ is the number of slots in the memory and $n_s$ is the dimension of each slot. Then, the network is formally parameterized as follows:
\begin{itemize}\vspace{-.3cm}
\item Initialize the top-layer factors $    \bbh_l^{(I_l)}  =  \bbz_{l+1}$.\vspace{-.3cm}
\item For $i = I_l - 1, \dots, 0$, do the transformation $\bbh_l^{(i)} = \phi(\bbh_l^{(i+1)}; M_{l}^{(i)})$,\vspace{-.3cm}
\end{itemize}
where $\phi$ is a proper (e.g., smooth) function for linear or nonlinear transformation. The bottom layer is our output $\bbmu_l = \bbh_l^0$, which is called a stochastic layer as it computes the mean parameters for a distribution to get samples from. All the other layers are called deterministic layers.

Compared with previous DGMs, one key feature of our model is that it incorporates an external memory at each deterministic layer, as detailed below. The overall architecture is a stack of multiple such layers interleaved with stochastic layers as above.
In such a DGM architecture, memory $M_l^{(i)}$ can recover the information that is missing in higher-layers $\bbh_l^{(>i)}$. In other words, the higher layers do not need to represent all details, but focusing on representing abstract invariant features if they seem more relevant to the task at hand than the more detailed information.

\subsection{General Memory Mechanism for a Single Layer}

We now present a single layer with memory generally, which is our building block for the above DGM. For notation simplicity, we omit the sub-script $l$ in the following text.
Formally, let $\bbh_{in}$ denote the input information, and $\bbh_{out}$ denote the output after some deterministic transformation with memory. In our model, $\bbh_{in}$ can be either the samples of latent factors or the output from a higher-level deterministic layer; and similarly $\bbh_{out}$ can be used as the input of either a stochastic layer or a lower-level deterministic layer.

A layer of standard DGMs without memory 
generates the low-level generative information $\bbh_{g}$ based on $\bbh_{in}$ through a proper transformation, which can be generally put as:
$$
\bbh_g = \phi(\bbh_{in}; W_g, b_g),
$$
where $W_g$ and $b_g$ are the weights and biases of the transformation respectively and uses it as the final output, i.e. $\bbh_{out} = \bbh_g$.

\begin{figure}[t]
\vskip 0.1in
\begin{center}
\includegraphics[width=.98\columnwidth]{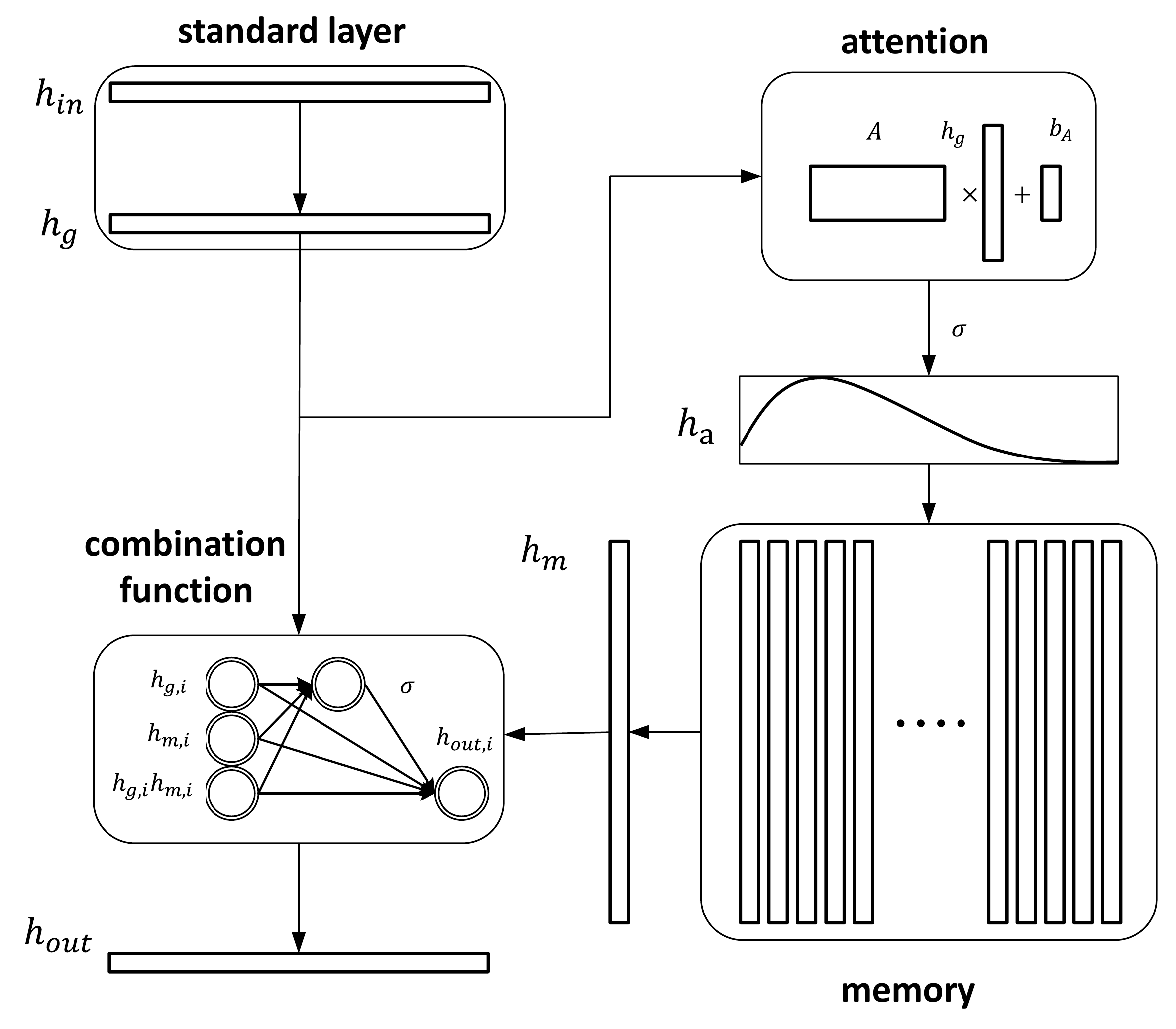}
\caption{Architecture comparison between a standard layer (top-left part) and a layer with memory (the whole figure).}
\label{arc_single}
\end{center}
\vskip -0.1in
\end{figure}

In our DGM with memory $M$, 
we first compute the low-level generative information $\bbh_g$ in the same way as a standard layer,
and then retrieve the memory with some proper attention mechanism to get knowledge $\bbh_m$.
Finally, we combine $\bbh_g$ and $\bbh_m$ to get the output $\bbh_{out}$. Formally, the memory retrieval process is parameterized as
$$
\bbh_m = f_m( \bbh_a; M),
$$
where $\bbh_a = f_a(\bbh_g;A, b_A)$ is the information used to access the memory and computed by an attention mechanism parameterized by a controlling matrix $A$ and a bias vector $b_A$.
The attention mechanism takes the generative information $\bbh_g$, which is the final output of a vanilla layer described previously, as input.
$f_a$ is the mapping function in the attention mechanism and $f_m$ is the mapping function in the memory mechanism, which are deterministic transformations to be specified.
The final output $\bbh_{out}$ is the combination of $\bbh_g$ and $\bbh_m$ as follows:
$$
\bbh_{out} = f_c (\bbh_g, \bbh_m; C),
$$
where $C$ is a set of trainable parameters in the combination function $f_c$, which
is another deterministic transformation to be specified.
We visualize the computation flow of these two types of layers in Figure~\ref{arc_single}, where each component will be specified next.

\subsection{Concrete Examples with Hierarchical Memory Mechanisms}
\label{sec:example-model}

With the above building blocks, we can stack multiple layers to build a DGM as in Section~\ref{sec:overall-struct}.
For simplicity, here we consider a generative model with only one stochastic layer and $I$ deterministic layers to explain our memory mechanism, which can be straightforwardly extended to cases with multiple stochastic layers.

Let the top most information to be the random samples from the prior, i.e., $\bbh^{(I+1)} = \bbz$.
%
%
Using permutation invariant architecture as an example, we compute the low-level generative information $\bbh_g^{(i)}$ based on the input $ \bbh^{(i+1)}$ as:
$$
\bbh_g^{(i)} = \phi(W_g^{(i)} \bbh^{(i+1)}+b_g^{(i)}).
$$
We further retrieve the knowledge $\bbh_m^{(i)}$ from memory. Though various strategies exist, we consider the simple one that adopts a linear combination of slots in memory as
$$
\bbh_m^{(i)} = f_m(\bbh_a^{(i)})= M^{(i)} \bbh_a^{(i)},
$$
where the coefficients $\bbh_a^{(i)}$ are computed as
$$\bbh_a^{(i)} = f_a(\bbh_g^{(i)}) = \sigma (A^{(i)} \bbh_g^{(i)} + b^{(i)}_A),$$
and $\sigma(x) = 1/ (1 + \textrm{exp}(-x))$ is the sigmoid function. Therefore, each element of $\bbh_a^{(i)}$ is a real value in the interval (0, 1),
which represents the preference of $\bbx$ to the corresponding memory slot.
An alternative soft attention function used in our experiment is the softmax function, which normalizes the summation of the preference values on all slots to be one~\cite{Bahdanau15}.
A hard attention mechanism trained with Reinforcement Learning~\cite{xu15} can be further investigated in the future work.

The most straightforward choice of the composition function is the element-wise summation:
$$
\bbh^{(i)} = \bbh_g^{(i)} + \bbh_m^{(i)},
$$
where the memory encodes the residual between the true target $\bbh^{(i)}$ and the generative information $\bbh^{(i)}_g$.
However, in practise, we found that a more flexible composition function can lead to a better result. Inspired by the Ladder Network~\cite{Valpola14,rasmus15},
we specify the combination function of $\bbh_m^{(i)}$ and $ \bbh_g^{(i)}$ as element wise multiple layer perceptron with optionally final nonlinearity $\phi$:
$$
\bbh^{(i)}= f_c(\bbh_g^{(i)}, \bbh_m^{(i)}) = \phi(a^{(i)} + b^{(i)}_1 c^{(i)}),
$$
where the inside linear part $a^{(i)}$ is the summation of scaled inputs and cross terms as well as biases:
\begin{eqnarray*}
a^{(i)} & = & a_1^{(i)} + a_2^{(i)} \odot \bbh_m^{(i)} + a_3^{(i)} \odot   \bbh_g^{(i)} \\ & + & a_4^{(i)} \odot \bbh_g^{(i)}  \odot \bbh_m^{(i)},
\end{eqnarray*}
and the inside nonlinear part $c^{(i)}$ is computed similarly but goes through a sigmoid function:
\begin{eqnarray*}
c^{(i)} &= & \sigma( c_1^{(i)} + c_2^{(i)} \odot \bbh_m^{(i)} + c_3^{(i)} \odot   \bbh_g^{(i)} \\ &+ &c_4^{(i)} \odot   \bbh_g^{(i)}  \odot \bbh_m^{(i)}),
\end{eqnarray*}
where  $\odot$ is the element wise product. The output in our model only depends on the top-down signals
$\bbh_g$ initially, instead of the auxiliary information as in the Ladder Network, which will be discussed in the experiment setting.
$(W_g^{(i)}, b_g^{(i)}, M^{(i)}, A^{(i)}, b_A^{(i)}, a_{1,2,3,4}^{(i)}, b_1^{(i)}, c_{1,2,3,4}^{(i)})$ are trainable parameters in single layer.
We illustrate each component in Figure.~\ref{arc_single}.

\section{Inference and Learning}

Learning a DGM is generally challenging due to the highly nonlinear transformations in multiple layers plus a stochastic formalism. To develop a variational approximation method, it is important to have a rich family of variational distributions that can well-characterize the nonlinear transforms. Significant progress has been made recently on stochastic variational inference methods with a sophisticated recognition model to parameterize the variational distributions~\cite{kingma14iclr,danilo14icml}. In this section, we develop such an algorithm for our DGM with memory.

Let $\bbtheta_g$ be the collection of parameters in the DGM. Then the joint distribution of each data $\bbx$ and the corresponding latent factor $\bbz$ can be generally put in a factorized form:
$$
p(\bbx, \bbz; \bbtheta_g) = p(\bbz ; \bbtheta_g) p(\bbx | \bbz;\bbtheta_g),
$$
where the prior is often of a simple form, such as spherical Gaussian in our experiments, and the form of the conditional distribution $p(\bbx | \bbz; \bbtheta_g)$ is chosen according to the data and its mean parameters depend on the external memories through a deep architecture as stated above.

\begin{figure}[t]
\vskip 0.1in
\begin{center}
\includegraphics[width=.98\columnwidth]{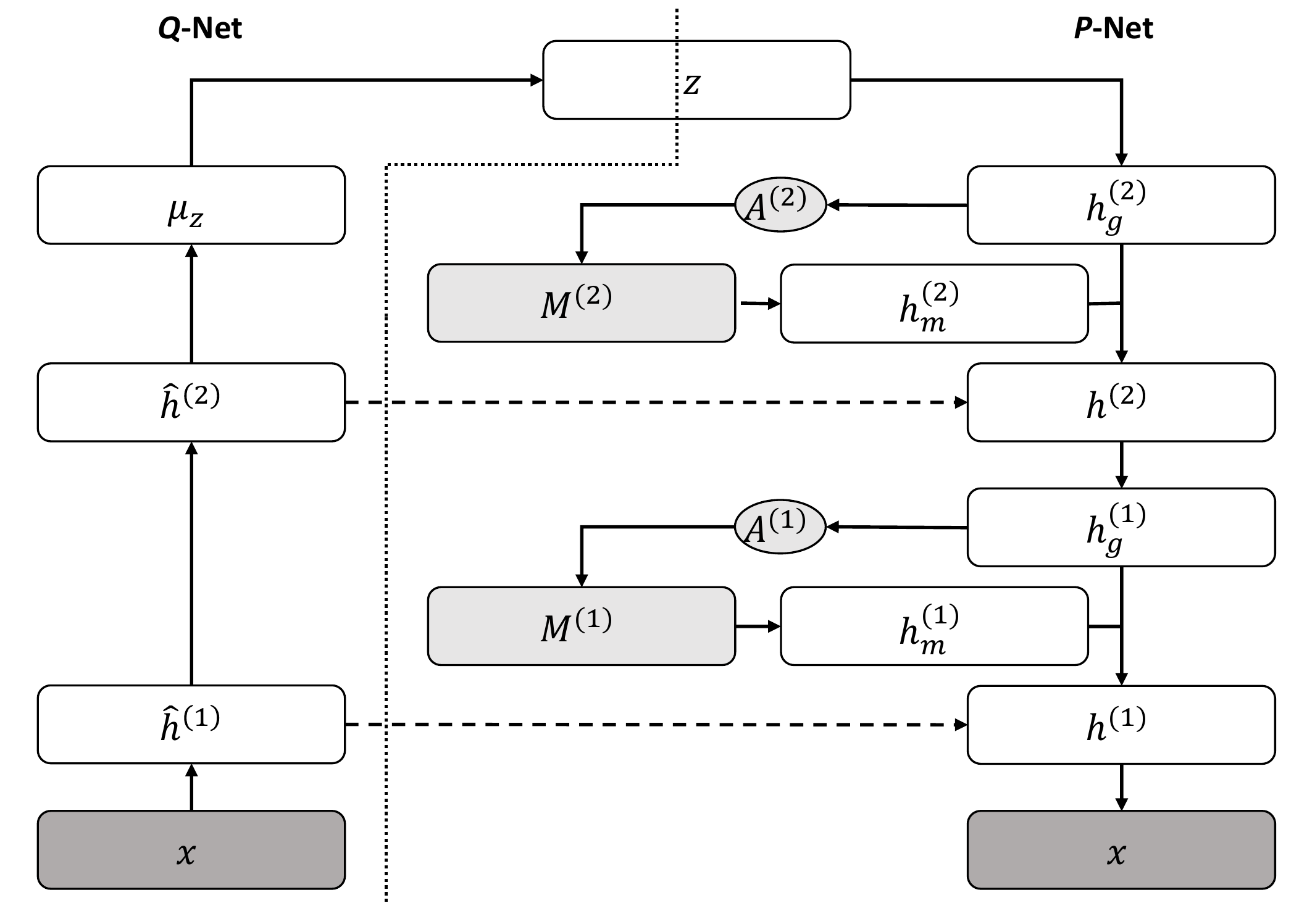}
\caption{ A model with one stochastic layer and two deterministic layers, where $\bbz$ is shared by the $P$-net and $Q$-net.}
\label{architecture}
\end{center}
\vskip -0.1in
\end{figure}

As in~\cite{kingma14iclr}, we adopt deep neural networks to parameterize a recognition model as the approximate posterior distribution $q(\bbz | \bbx; \bbtheta_r)$, where $\bbtheta_r$ is the collection of the parameters in the recognition model (denoted by $Q$-Net, as it characterizes distribution $q$).
Since the $Q$-Net implements the bottom-up abstraction process to identify invariant features, it is unnecessary to have an external memory. Furthermore, the $Q$-Net without memory is compact in parameterization.
The overall architecture is asymmetric, as illustrated in Figure~\ref{architecture}, where the components at the left side of the dot line together with sampling $\bbz$ from $q(\bbz | \bbx)$ is the $Q$-Net and the components at the right side of the dot line with $\bbz$ sampled from the prior is the generative model (denoted by $P$-Net, as it characterizes model distribution $p$).
The solid arrow means the corresponding component is used as input of next component and the dash arrow means the corresponding component is used as the training target of next component, as explained below.
The components representing external memory and associated attention mechanisms are filled with shallow gray.
We omit the components corresponding to the combination functions for better visualization.


We define the $Q$-Net as follows. Following the example with one stochastic layer and $I$ deterministic layers in the previous section, we extract the high-level features $\hat \bbh^{(i + 1)}$ as follows:
$$
\hat \bbh^{(i + 1)} =  \phi(V^{(i)} \hat \bbh^{(i)}+b_r^{(i)}),
$$
where $\phi$ is a proper nonlinear function and $(V^{(i)}, b_r^{(i)})$ are trainable parameters.
The bottom layer is the input data, i.e. $\hat \bbh^{(0)} = \bbx$ and the top layer is still factorized Gaussian distribution. The mean of $\bbz$ is computed by linear transformation of $\hat \bbh^{(I)}$ and the variance of $\bbz$ is computed similarly but with a final exponential nonlinearity.

A variational lower bound of log-likelihood for per data $\bbx$ can be formulated as:
$$
\mathcal{L} (\bbtheta_g, \bbtheta_r; \bbx) \triangleq \mathbb{E}_{q(\bbz | \bbx; \bbtheta_r)}[\log p(\bbx, \bbz; \bbtheta_g) - \log q(\bbz | \bbx; \bbtheta_r)].
$$

We add local reconstruction error terms as an optional regularizer, and jointly optimize the parameters in the generative model and the recognition model:
$$
 \min_{\bbtheta_g, \bbtheta_r}  \frac{1}{|\mathcal{D}|} \sum_{\bbx \in \mathcal{D}} \big(\mathcal{L}(\bbtheta_g, \bbtheta_r; \mathbf{x})
 +  \sum_{i = 1}^{I} \lambda^{(i)} || \bbh^{(i)} - \hat \bbh^{(i)}||_2^2 \big),
$$
where the relative weights $\lambda^{(i)}$ are prefixed hyperparameters.
We optimize the objective with a stochastic gradient variational Bayes (SGVB) method~\cite{kingma14iclr}.
Note that we cannot send the message of a intermediate layer in the recognition model to a layer in the generative model through a lateral connection as in Ladder Network~\cite{Valpola14,rasmus15} because that indeed changes the distribution of $p(\bbx|\bbz)$ according to the data $\bbx$.
However, we do not use any information of $\bbx$ in the generative model explicitly and
the correctness of the variational bound can be verified.

We employ batch normalization layers~\cite{ioffe15} in both the recognition model and generative model
to accelerate the training procedure, and the intermediate features in local reconstruction error terms are replaced by a corresponding normalized version. To compare with stat-of-the-art results, we also train our method as in importance weighted autoencoders (IWAE)~\cite{burda15}, which uses importance weighting estimate of log likelihood with multiple samples in the training procedure to achieve a strictly tighter variational lower bound.


\section{Experiments}
\label{body-exp}

We now present both quantitative and qualitative evaluations of our method on the real-valued MNIST, OCR-letters and Frey faces datasets for various tasks.
The MNIST dataset~\cite{Lecun:98} consists of 50,000 training, 10,000 validation and 10,000 testing images of handwritten digits and
each image is of $28\times 28$ pixels.
The OCR-letters dataset~\cite{Bache13} consists of
32,152 training, 10,000 validation and 10,000 testing letter images of size $16 \times 8$ pixels. The Frey faces dataset consists of
1,965 real facial expression images of size $28 \times 20$ pixels.
We model MNIST and OCR-letters datasets as Bernoulli distribution and model Frey faces dataset as Gaussian distribution at data level.

Our basic competitors are VAE~\cite{kingma14iclr} and IWAE~\cite{burda15}. We add the memory mechanisms to these methods and denote our models as MEM-VAE and MEM-IWAE, respectively. In all experiments except the visualization in Appendix \ref{app-vis}, MEM-VAE employs
the sigmoid function and element-wise MLP as the attention and composition functions respectively.

Our implementation is based on Theano~\cite{bastein2012}.\footnote{Source code at https://github.com/zhenxuan00/MEM\_DGM}
We use ADAM~\cite{Kingma:15} in all experiments with parameters $\beta_1 = 0.9$, $\beta_2 = 0.999$ (decay rates of moving averages) and $\epsilon=10^{-4}$ (a constant that prevents overflow). As a default, the global learning rate is fixed as $10^{-3}$ for 1,000 epochs and annealed by a factor 0.998 for 2,000 epochs with minibatch size 100.
Initially, We set $a_3^i$ and $c_3^i$ as vectors filled with ones and $(a_{1,2,4}^{(i)}, b_1^{(i)}, c_{1,2,4}^{(i)})$ as vectors filled with zeros to avoid poor local optima. This means that we initialize the output as signals from top-down inference, which is different from the Ladder Network~\cite{rasmus15}. We initialize the memory matrix as Gaussian random variables and other parameters following~\cite{glorot15}. We specify $\phi$ as rectified linear units (ReLu)~\cite{nair10} in both the generative model and the recognition model.

We do not tune the hyper-parameters of our method heavily. We choose a model with one stochastic layer and two deterministic layers as the default setting.
The values of $\lambda^{(1)}$ and $\lambda^{(2)}$ are fixed as $0.1$ following Ladder Network~\cite{rasmus15}.
We do not include a local reconstruction error term at data level since the variational lower bound penalizes the reconstruction error of data already.
The dimension of slots in memory $d_s$ is the same as that of the corresponding generative information $\bbh_g$ because we use element-wise combination function $f_c$. We employ the memory mechanism in both deterministic layers and make the total number of slots $n_s^{(1)} + n_s^{(2)}$ to be 100 to keep the number of additional parameters relatively small. We choose a 70-30 architecture according to the validation performance on the MNIST dataset and then make it default for all experiments if not mentioned.

\subsection{Density Estimation}
\label{sec_density_estimation}

We follow~\cite{burda15} to split the MNIST dataset into 60,000 training data and 10,000 testing data after choosing the hyper-parameters.
We train both the baselines and our models with 1, 5 and 50 importance samples respectively and evaluate the test likelihood with 5,000 importance samples as in~\cite{burda15}.
In each training epoch, we binarize the data stochastically as the input.
The results of VAE, IWAE-5 (trained with 5 importance samples) and IWAE-50 (trained with 50 importance samples) with one stochastic layer in~\cite{burda15} are -86.76, -85.54 and -84.78 nats respectively.
However, we use 500 hidden units in the deterministic layers and 100 latent variables in the stochastic layer to achieve a stronger baseline result with a different architecture and more parameters. We present our likelihood results in Table~\ref{basic-table}. We can see that our methods improve the results of baselines (both VAE and IWAE) significantly and achieve state-of-the-art results on the real-valued MNIST dataset with permutation invariant architectures.
DRAW~\cite{gregor15} achieves -80.97 nats by exploiting the spatial information.
Our method MEM-IWAE-50 even outperforms {\it S2-IWAE-50}, which is the best model in~\cite{burda15} with two stochastic layers and four deterministic layers.

To compare with a broader family of benchmarks, we further quantitatively evaluate our model on the OCR-letters dataset.
We use 200 hidden units in the deterministic layers and 50 latent variables in the stochastic layer as the dimension of the input is much smaller. The test log-likelihood is evaluated with 100,000 importance samples as in~\cite{bornschein14} and shown in Table~\ref{basic-table}. Again, our methods outperform the baseline approaches significantly and are comparable with the best competitors, which often employ autoregressive connections~\cite{Larochelle:11,Gregor:14} that are effective on small images with simple structures.
Note that these sophisticated structures are not exclusive to our memory mechanisms. A systematic investigation of using memory with such structures is our future work.

\begin{table}[t]
\caption{Log likelihood estimation on MNIST and OCR-letters datasets. Results are from  [1]~\cite{murray09}, [2]~\cite{burda15}, [3]~\cite{bornschein14}, [4]~\cite{Larochelle:11} and [5]~\cite{Gregor:14}.
Results with * are evaluated on binarised MNIST dataset.}
\label{basic-table}
\vskip 0.15in
\begin{center}
\begin{small}
\begin{sc}
\begin{tabular}{lcc}
\hline
\abovespace\belowspace
Models & MNIST & OCR-letters \\
\hline
\abovespace
{\it VAE} & -85.67 & -30.09\\
{\it MEM-VAE(ours)} & -84.41 & -29.09\\
\hline
\abovespace
{\it IWAE-5} & -84.49 & -28.69\\
{\it MEM-IWAE-5(ours)} & -83.26 & -27.65\\
\hline
\abovespace
{\it IWAE-50} & -83.67 & -27.60\\
{\it MEM-IWAE-50(ours)} & -\textbf{82.84} & -\textbf{26.90}\\
\hline
\abovespace
{\it DBN}[1] & -84.55 & - \\
{\it S2-IWAE-50}[2] & -\textbf{82.90} & -\\
\hline
\abovespace
{\it RWS-SBN/SBN}[3]* & -85.48 & -29.99\\
{\it RWS-NADE/NADE}[3]* & -85.23 & -\textbf{26.43}\\
{\it NADE}[4]* & -88.86 & -27.22 \\
{\it DARN}[5]* & -\textbf{84.13} & -28.17 \\
\hline
\end{tabular}
\end{sc}
\end{small}
\end{center}
\vskip -0.15in
\end{table}

\subsection{Analysis of Our Model}

We now present a careful analysis of our model to investigate the possible reasons for the outstanding performance.

\begin{figure*}[t]
\vskip 0.05in
\centering
\subfigure[Layer 2]{\includegraphics[height=0.2\columnwidth,width=0.36\columnwidth]{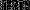}}
\subfigure[Layer 1]{\includegraphics[height=0.2\columnwidth,width=0.84\columnwidth]{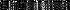}}
\subfigure[Layer 2]{\includegraphics[width=0.39\columnwidth]{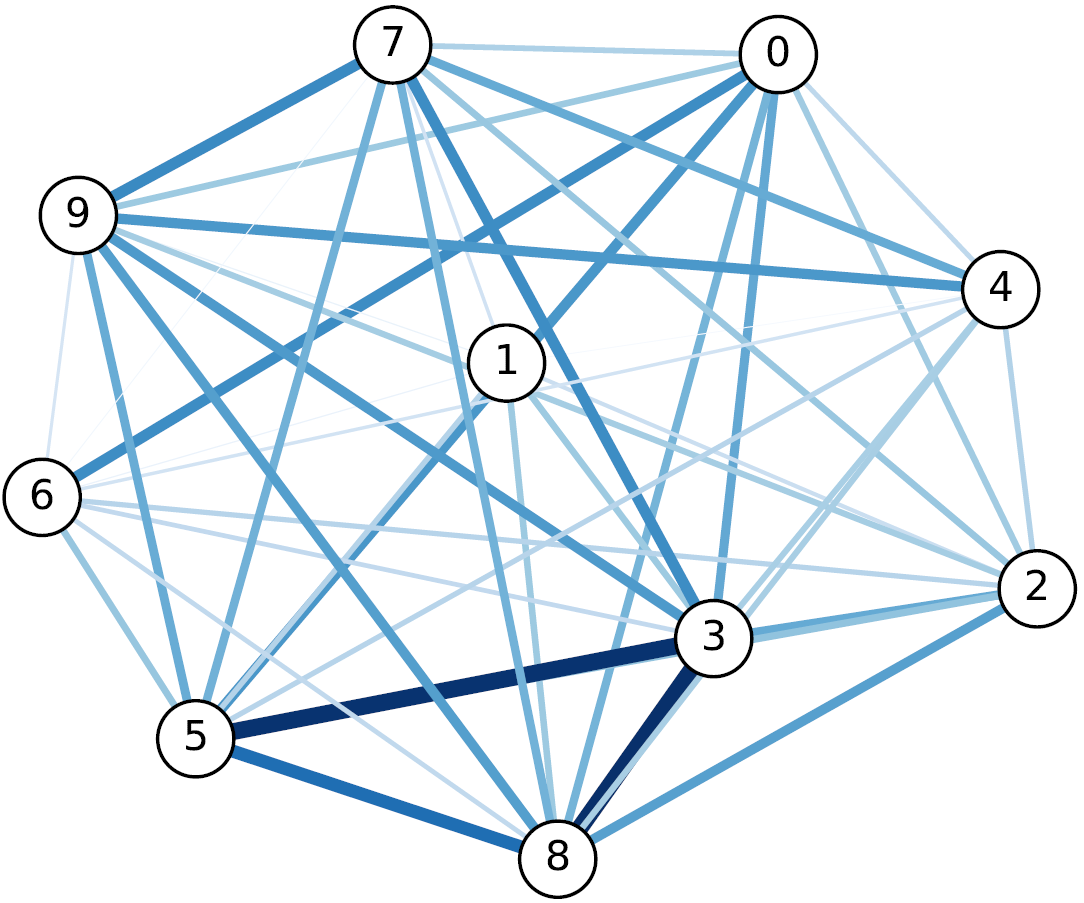}}
\subfigure[Layer 1]{\includegraphics[width=0.39\columnwidth]{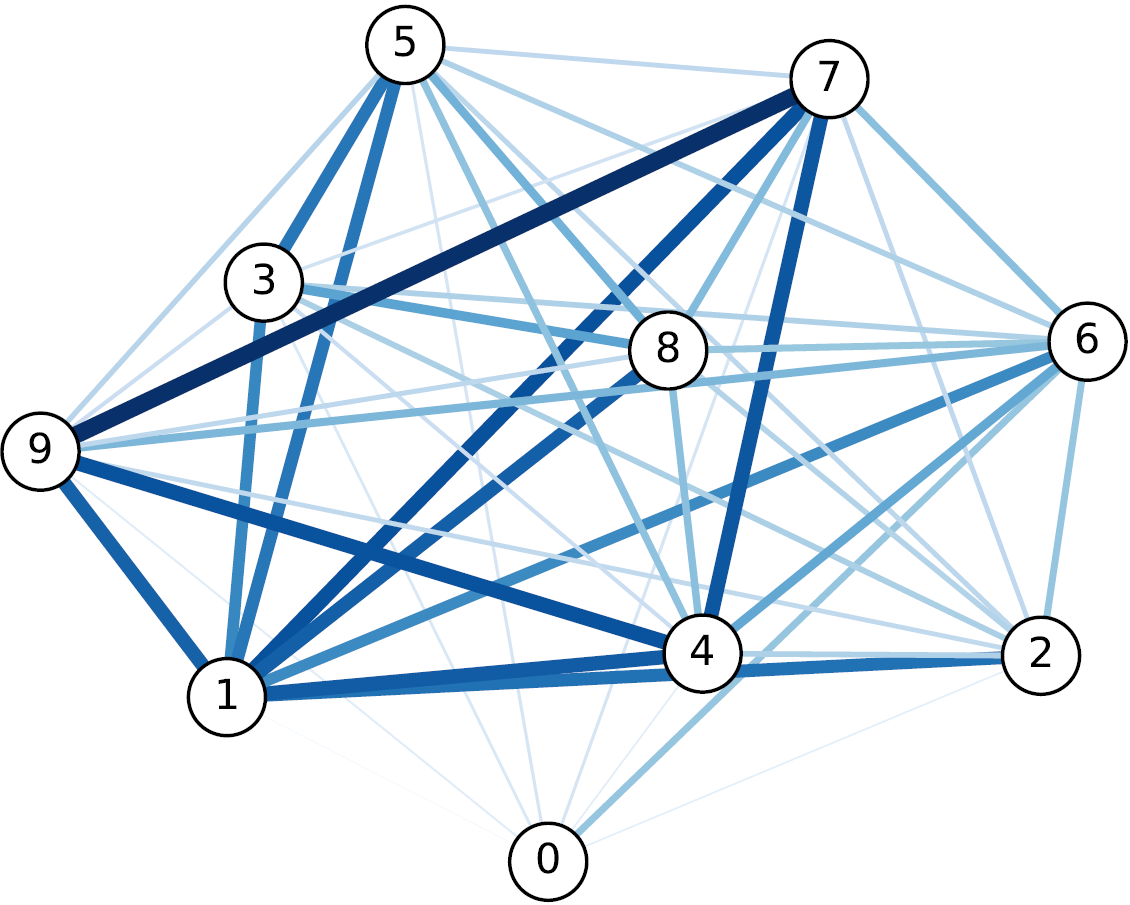}}
\caption{(a-b): Averaged activations on each memory slot over different classes of testing data on MNIST dataset in layer 2 and layer 1 respectively.
(c-d): 2-D visualization for correlation between classes for layer 2 and layer 1 respectively (best view in color).}
\label{memory_visualization}
\vskip -0.18in
\end{figure*}

\textbf{Classification:} We investigate the effect of external memory on the training of the recognition model by classification
 and MEM-VAE outperforms VAE (See details in Appendix~\ref{app-rec}).

\textbf{A larger baseline:}
We test VAE with a 530-530-100 architecture, which has almost the same number of parameters as MEM-VAE. The log-likelihood trained with 1, 5 and 50 importance samples on MNIST are -85.69, -84.43 and -83.58 respectively. We can see that using our memory leads to much better results than simply increasing the model size. A comparison of number of parameters used in all of the models can be seen in Appendix~\ref{app-numofp}.

\textbf{Importance of memory:}
We test the relative importance of the memory mechanism and local reconstruction error regularizer.
MEM-VAE in the default settings but without local reconstruction error regularizer achieves a test log density estimation of -84.44 nats.
VAE with additional local reconstruction regularizer achieves test log density estimation of -85.68 nats.
These experiments demonstrate that the memory mechanism plays a central role in the recovery of detailed information.
The local reconstruction error regularizer may help more provided supervision.

\textbf{Preference of memory slots over classes:}
We investigate the preference of memory slots over different classes in MEM-VAE.
We average $\bbh_a$ and normalize the activations for each class and visualize the matrices in Figure~\ref{memory_visualization}(a-b),
where each column represents a slot and each row represents a class (0-9 in top-down order).
The averaged and normalized activations are used as the intensities for the corresponding positions in the matrices.
Furthermore, we compute the correlation coefficients between activations of different classes and visualize them in a 2-D graph in Figure~\ref{memory_visualization}(c-d),
where each node represents a class and each edge represents the correlation between two endpoints. The larger the correlation is, the wider and darker the edge is.
We observe that the trained attention model can access the memory based on the implicit label information in the input, which accords with our assumption. The activations are correlated for those digits that share similar structures such as ``7'' and ``9''. Furthermore, different layers of memory focus on different patterns. For example, layer 1 has a strong activation of a vertical line pattern which is shared among digits ``1'', ``4'', ``7'' and ``9'', while layer 2 activates most to a semi-circle pattern which is shared among digits ``3'', ``5'' and ``8''. Besides, layer 1 has almost the same 2D-visualization result as the raw data.

\textbf{Visualization:} We visualize the generative information $\bbh_g$ and memory information $\bbh_m$ by mapping these vectors to images (See details in Appendix~\ref{app-dis} and~\ref{app-vis} respectively).

\subsection{Random Generation}

\begin{figure}[t]
\centering
\subfigure[IWAE-50]{\includegraphics[width=0.49\columnwidth]{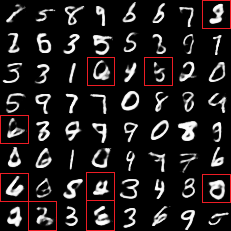}}
\subfigure[MEM-IWAE-50]{\includegraphics[width=0.49\columnwidth]{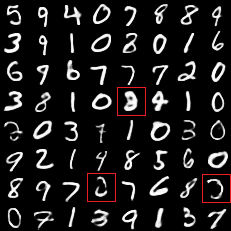}}
\caption{(a-b): Random generation from IWAE-50 and MEM-IWAE-50 on MNIST dataset respectively.}
\label{mlp_mnist_sample}
\vskip -0.15in
\end{figure}

\begin{figure}[t]
\centering
\subfigure[VAE]{\includegraphics[width=0.49\columnwidth]{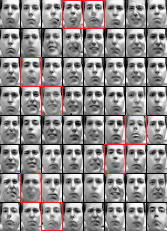}}
\subfigure[MEM-VAE]{\includegraphics[width=0.49\columnwidth]{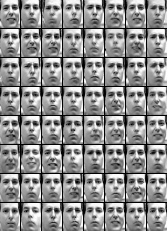}}
\caption{(a-b): Random generation from VAE and MEM-VAE on Frey faces dataset respectively.}
\label{mlp_frayface_sample}
\end{figure}

We further evaluate the random generations from the baseline and our model empirically on MNIST and Frey faces datasets,
which is shown in Figure~\ref{mlp_mnist_sample} and Figure~\ref{mlp_frayface_sample} respectively.
We label unclear or meaningless images with red rectangles. This is done by majority voting of several volunteers.
We do not select any pictures for both datasets.

For the MNIST dataset, the setting is same as in Section~\ref{sec_density_estimation}. We observe that the memory mechanism helps a lot to get clear and meaningful samples as in Figure~\ref{mlp_mnist_sample}.

For Frey faces dataset, we randomly split into 1,865 training data and 100 testing data.
We use a single deterministic layer with 200 hidden units and a stochastic layer with 10 latent factors and set $n_s^{(1)}$ to be 20 as the number of training samples is small. We use one sample of the recognition model in both of the training and testing procedure as in~\cite{kingma14iclr}.
We find that the minibatch size effects the results a lot, and the quality of visualization and the averaged test log density are inconsistent~\cite{theis16}. Specifically,
setting the minibatch size to be 100, VAE achieves test log density of 1308 nats, which reproduces the result with same architectures in~\cite{kingma14iclr}, but the visualization is somehow unclear; while
setting the minibatch size to be 10, VAE achieves test log density of 1055 nats, but the visualization is much better.
All of the parameters are set referred to~\cite{kingma14iclr} or based on the performance of test log density of VAE.
We also find that MEM-VAE outperforms VAE in both cases in terms of the quantitative test likelihood and qualitative visualization --- the corresponding log density of MEM-VAE are 1330 and 1240 nats respectively.
The random samples given minibatch size 100 is shown in Figure~\ref{mlp_frayface_sample}, where we can see that all samples of MEM-VAE are clear but some of VAE cannot present all details in the facial expression successfully.

\subsection{Missing Value Imputation}

Finally, we evaluate our method on the task of missing value imputation with three different types of noise, including
(1) \emph{RECT}-12 means that a centered rectangle of size $12 \times 12$ is missing; (2) \emph{RAND-}0.6 means that each pixel is missing with a prefixed probability $0.6$;
and (3) \emph{HALF} means that the left half of the image is missing.
For both VAE and MEM-VAE, the missing values are randomly initialized and then inferred by a Markov chain that samples latent factors based on the current guess of missing values and then refines the missing values based on the current latent factors.
We compare the mean square error (MSE) results after 100 epochs of inference as in Table~\ref{mse-mnist} on the MNIST dataset. The results demonstrate that DGM with external memory can capture the underlying structures of data better than vanilla methods under different types of noise.
Besides, MEM-VAE has better qualitative results (See Appendix~\ref{app-mvi}).

\begin{table}[t]
\caption{MSE results on MNIST dataset with different types of noise.}
\label{mse-mnist}
\begin{center}
\begin{small}
\begin{sc}
\begin{tabular}{lccr}
\hline
\abovespace\belowspace
Noise type & VAE & MEM-VAE \\
\hline
\abovespace
{\it RECT}-12 & 0.1403 & {\bf 0.1362}\\
{\it RAND}-0.6 & 0.0194 & {\bf 0.0187} \\
{\it HALF} & 0.0550 & {\bf 0.0539}\\
\hline
\end{tabular}
\end{sc}
\end{small}
\end{center}
\vskip -0.2in
\end{table}

\section{Conclusions and Future Work}

In this paper, we introduce a novel building block for deep generative models (DGMs) with an external memory and an associated soft attention mechanism. In the top-down generative procedure, the additional memory helps to recover the local detail information, which is often lost in the bottom-up abstraction procedure for learning invariant representations.
Various experiments on handwritten digits and letters as well as real faces datasets demonstrate that our method can substantially improve the vanilla DGM on density estimation, random generation and missing value imputation tasks, and we can achieve state-of-the-art results among a broad family of benchmarks.

There are three possible extensions of our method:
\begin{itemize}\vspace{-.3cm}
  \item The use of other types of memory and attention mechanisms in DGMs can be further investigated. Particularly, the combination of external memory and visual attention as well as recurrent networks~\cite{gregor15} may achieve better results in generative tasks.\vspace{-.2cm}
  \item A class conditional DGM~\cite{kingma14nips} with memory can potentially achieve better performance on both classification and generation because the external memory helps to reduce the competition between the invariant feature extraction and detailed generation, and explicit label information can make the whole system be easier to train.\vspace{-.2cm}
  \item Our method can be further applied to convolutional neural networks by sharing parameters across different channels and
  then employed in non-probabilistic DGMs such as LAPGAN~\cite{denton15} to refine generation on high-dimensional data. \vspace{-.2cm}
\end{itemize}

\section*{Acknowledgments}

The work was supported by the National Basic Research Program (973 Program) of China (Nos. 2013CB329403, 2012CB316301), National NSF of China (Nos. 61322308, 61332007), the Youngth Top-notch Talent Support Program, Tsinghua TNList Lab Big Data Initiative, and Tsinghua Initiative Scientific Research Program (No. 20141080934).

\bibliography{example_paper}
\bibliographystyle{icml2016}

\clearpage

\appendix

\section{Recognition Model}
\label{app-rec}

We feed the output of the last deterministic layer in the recognition models into a linear SVM to classify the MNIST digits to examine the invariance in features.
We achieve slightly better classification accuracy (97.90\% for VAE and 98.03\% MEM-VAE), which means that additional memory mechanisms do not hurt or even improve the invariance of the features extracted by the recognition model.

\section{Number of Parameters Used}
\label{app-numofp}
As we employ external memory and attention mechanisms, the number of parameters in our building block is larger than that in a standard layer.
However, the total number of parameters in the whole model is controlled given a limited number of slots in the memory. See Table~\ref{paras-practise} for a comparison, and we do not observe that our method suffers from overfitting.

\begin{table}[t]
\caption{Number of parameters used in practice.}
\label{paras-practise}
\begin{center}
\begin{small}
\begin{sc}
\begin{tabular}{lccr}
\hline
\abovespace\belowspace
Model & Number of parameters \\
\hline
\abovespace
{\it MNIST-500-500-100} & 1,441k\\
{\it MNIST-530-530-100} & 1,559k\\
{\it MNIST-500-500-100-MEM} & 1,550k\\
\hline
\abovespace
{\it OCR-letters-200-200-50} & 164k\\
{\it OCR-letters-200-200-50-MEM} & 208k\\
\hline
\end{tabular}
\end{sc}
\end{small}
\end{center}
\end{table}

\section{Disabling Memory}
\label{app-dis}
\begin{figure}
\centering
\includegraphics[width=.98\columnwidth]{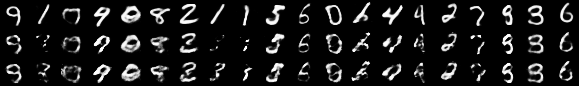}
\caption{Generation from MEM-VAE when memory is disabled.}
\label{disable_visualization}
\end{figure}

We investigate the performance of MEM-VAE when the memory is disabled (setting $\bbh_m$ as a vector filled with ones) as in Figure~\ref{disable_visualization}.
The top row shows original samples; the middle row shows samples with memory of the first layer disabled;
and the bottom row shows samples with memory of both layers disabled. It can be seen that, without information from memory, the main pattern of the generation does not change much but the local details are lost in some sense, which supports our assumption.

\section{Visualizing the Memory Slots Directly}
\label{app-vis}
As mentioned in Section \ref{body-exp}, MEM-VAE employs the sigmoid function and element-wise MLP as the attention and composition functions respectively.
MEM-VAE is complicated and flexible, which has better quantitative results on log-likelihood estimation but is hard to be visualized directly as it has much nonlinearity.
We introduce VIS-MEM-VAE, which uses the softmax function and element-wise summation as the attention and composition functions respectively. VIS-MEM-VAE achieves a slightly worse density estimation result (-84.68 nats on MNIST, still better than -85.67 nats of VAE) but the memory slots can be mapped to data level for visualization due to the simple composition function and sparse attention mechanism.

\begin{figure}[t]
\begin{center}
\includegraphics[width=.98\columnwidth]{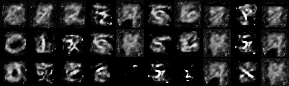}
\caption{Visualization of selected slots.}
\label{s-vis}
\end{center}
\end{figure}

We average the preference values $\bbh_a$ of the test data and select top-3 memory slots that has the highest activations for each class.
Note that the activations are normalized, i.e., the summation of the preference values on all memory slots equals to one for each class.
We set the generative information to be a vector filled with zeros and the memory information to be one of the selected slots and then generate a corresponding image. The activations and images are shown in Table~\ref{s-acti} and Figure~\ref{s-vis}, where
each column represents a class (0-9 in left-right order).
For example, the image at the first column and the second row in Figure~\ref{s-vis} corresponds to the memory slot that has the second-highest averaged activation of class ``0'' and the value is 0.24.

It can be seen that most of the selected slots respond to one class or similar classes (some slots are shared by similar classes such as``4'', ``7'' and ``9'') and the corresponding image contains a blurry sketch of the digit (or mixture of some digits) with different local styles, which indicates that the external memories can encode local variants of objects and can be retrieved based on generative information $\bbh_g$.
A few images are less meaningful but the corresponding activations are relatively small (smaller than 0.08 or so).

\begin{table}[t]
\caption{Average preference values of selected slots.}
\label{s-acti}
\begin{center}
\vskip 0.145in
\begin{small}
\begin{sc}
\begin{tabular}{p{.35cm}p{.35cm}p{.35cm}p{.35cm}p{.35cm}p{.35cm}p{.35cm}p{.35cm}p{.35cm}p{.35cm}}
\hline
\abovespace\belowspace
``0'' & ``1'' & ``2'' & ``3'' & ``4'' & ``5'' & ``6'' & ``7'' & ``8'' & ``9''\\
\hline
\abovespace
0.27 & 0.82 & 0.33 & 0.11 & 0.34 & 0.15 & 0.49 & 0.27 & 0.09 & 0.28\\
0.24 & 0.09 & 0.06 & 0.11 & 0.30 & 0.13 & 0.12 & 0.27 & 0.09 & 0.21\\
\belowspace
0.18 & 0.05 & 0.06 & 0.11 & 0.07 & 0.07 & 0.05 & 0.11 & 0.09 & 0.18\\
\hline
\end{tabular}
\end{sc}
\end{small}
\end{center}
\end{table}

\section{Missing Value Imputation}
\label{app-mvi}

\begin{figure}[t]
\centering
\subfigure[Data]{\includegraphics[width=0.48\columnwidth]{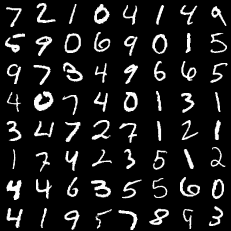}}
\subfigure[Noisy data]{\includegraphics[width=0.48\columnwidth]{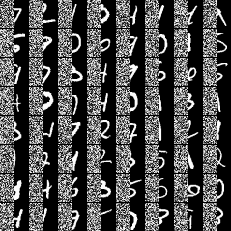}} \\
\subfigure[Results of VAE]{\includegraphics[width=0.48\columnwidth]{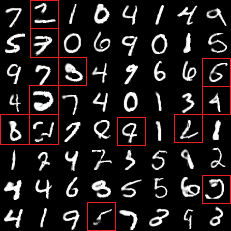}}
\subfigure[Results of MEM-VAE]{\includegraphics[width=0.48\columnwidth]{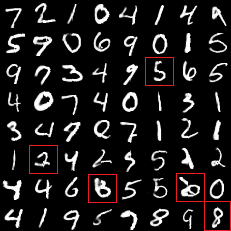}}
\caption{(a-b): Original test data on MNIST and perturbed data with left half missing respectively;
(c-d): Imputation results of VAE and MEM-VAE respectively. Red rectangles mean that the corresponding model fails to infer digits correctly but the competitor succeeds.}
\label{denoise_half}
\end{figure}

We visualize the images recovered by VAE and MEM-VAE in Figure~\ref{denoise_half} given incomplete test data.
It can be seen that MEM-VAE has better visualization than VAE with fewer meaningless images, clearer digits and more accurate inference.

\end{document}